\documentclass[letterpaper]{article} 
\usepackage{aaai2027}  
\usepackage[hyphens]{url}  
\usepackage{graphicx} 
\usepackage{natbib}  
\usepackage{caption} 
\usepackage{algorithm}
\usepackage{algorithmic}

\usepackage{newfloat}
\usepackage{listings}
\DeclareCaptionStyle{ruled}{labelfont=normalfont,labelsep=colon,strut=off} 
\floatstyle{ruled}
\newfloat{listing}{tb}{lst}{}
\floatname{listing}{Listing}

\usepackage{booktabs}
\usepackage{amssymb}

\nocopyright

\title{Per-View Gaussian Predictions Enable Training-Free Distractor Filtering \\in Feed-Forward 3DGS}
\author{
    Kangmin Seo, 
    Jae-Pil Heo\corresponding
}
\affiliations{
    Sungkyunkwan University

    skmskku@skku.edu, jaepilheo@skku.edu
}

\begin{document}

\maketitle

\begin{abstract}
Feed-forward 3D Gaussian Splatting reconstructs an explicit Gaussian
representation from multiple input images in one network execution,
making 3D reconstruction increasingly accessible for casual captures.
However, such captures frequently contain transient objects that
appear in only a subset of the views. Such content can be encoded into the per-view Gaussians associated with the inputs that observe it and remain in the combined representation despite being observed by no other input. As a result, it may produce blurred, duplicated, or
floating artifacts in novel views.
We introduce a training-free filtering procedure that exploits this
per-view prediction structure. For each input, we exclude its associated Gaussians and render the same camera using the remaining representation, revealing content that is inconsistent with the other inputs. Feature similarity forms candidate regions, and rendering-based
verification retains only candidates whose removal reduces
reconstruction error in the other input views. The procedure operates
on a single frozen prediction without retraining or scene-specific
optimization. Across three reconstruction models and two distractor
benchmarks, it consistently improves novel-view quality with varying
numbers of input views. On clean scenes, evaluations across four models show that the original reconstructions are largely preserved.
\end{abstract}
\section{Introduction}

Recent advances in neural radiance fields~\cite{nerf} and
3D Gaussian Splatting~\cite{3dgs} have substantially improved
novel view synthesis. More recently, feed-forward 3DGS models
directly predict a Gaussian representation from a set of input images
in one network execution
~\cite{pixelsplat,mvsplat,depthsplat,noposplat,anysplat},
enabling fast and generalizable reconstruction.

These methods commonly assume that the input images depict a static
and view-consistent scene. Casual captures, however, often contain
pedestrians, vehicles, and other transient objects that appear in only
one or a small subset of the images. When incorporated into the
predicted representation, such distractors can produce blurry regions,
duplicated appearance, and floating artifacts in novel views.

\begin{figure}[t]
    \centering
    {\fboxsep=0pt\fboxrule=0pt\fbox{\includegraphics[width=\dimexpr\columnwidth-2\fboxrule\relax]{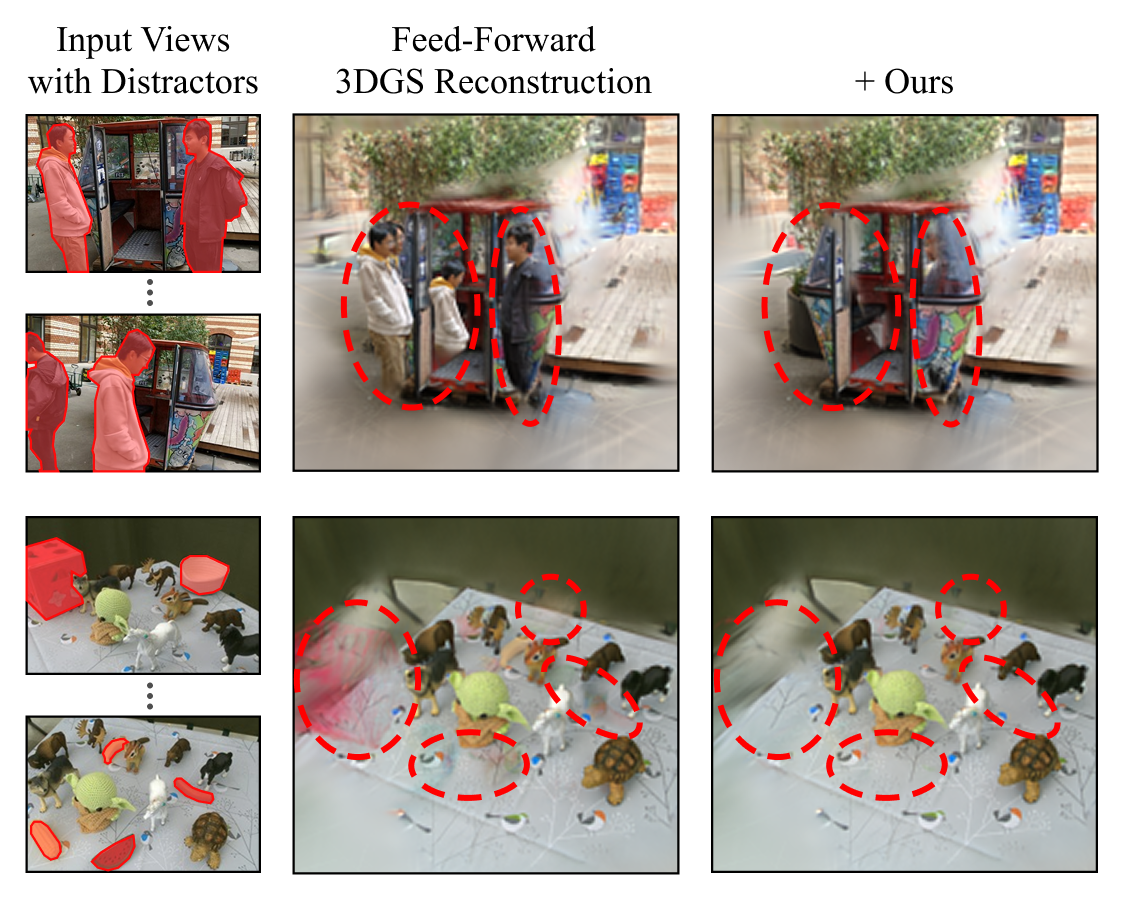}}}
\caption{Qualitative overview. Distractors in the input views produce
artifacts in the feed-forward reconstruction, which our method
suppresses on the same frozen prediction.}
    \label{fig:teaser}
\end{figure}

Distractor handling has primarily been studied in optimization-based
NeRF and 3DGS pipelines. Existing methods model transient appearance
~\cite{nerf-w}, use robust objectives or uncertainty estimation
~\cite{robustnerf,nerf-on-the-go}, leverage pretrained features or
residual cues~\cite{wildgaussians,spotlesssplats}, separate static and
transient content~\cite{desplat,hybridgs,forestsplats}, or use
consistency and multistage filtering
~\cite{asymgs,pdf-gs,dualsplat}. While fitting a scene to all input
images, differences between its renderings and individual observations
can provide useful evidence of unsupported content.

With the emergence of feed-forward reconstruction, recent work has
introduced distractor-aware variants through dedicated training,
learned mask prediction, reconstruction input selection, or Gaussian
pruning~\cite{dggs,genwildsplat,vgtw}. These approaches address distractors by adapting particular reconstruction backbones, relying on distractor-specific trained components such as mask prediction heads or semantic segmentation priors. Consequently, transferring distractor robustness to a different reconstruction backbone may require model-specific adaptation. In contrast, we ask whether such robustness can be obtained from the reconstruction itself, without modifying the model.

An immediate candidate for such a cue is the rendering-input discrepancy exploited in optimization-based pipelines. In the feed-forward setting, however, this cue is obscured. The reconstruction models considered in this work retain an association between predicted Gaussians and their input views~\cite{depthsplat,resplat,yonosplat,anysplat}. A model can therefore preserve content from an individual input even when that content is inconsistent with the remaining observations. The associated Gaussians
may reproduce it when rendered from the corresponding camera,
concealing the discrepancy with the input image. The same Gaussians
nevertheless remain in the shared 3D representation and may produce
artifacts from other viewpoints.

\begin{figure}[t]
    \centering
    {\fboxsep=0pt\fboxrule=0pt\fbox{\includegraphics[width=\dimexpr\columnwidth-2\fboxrule\relax]{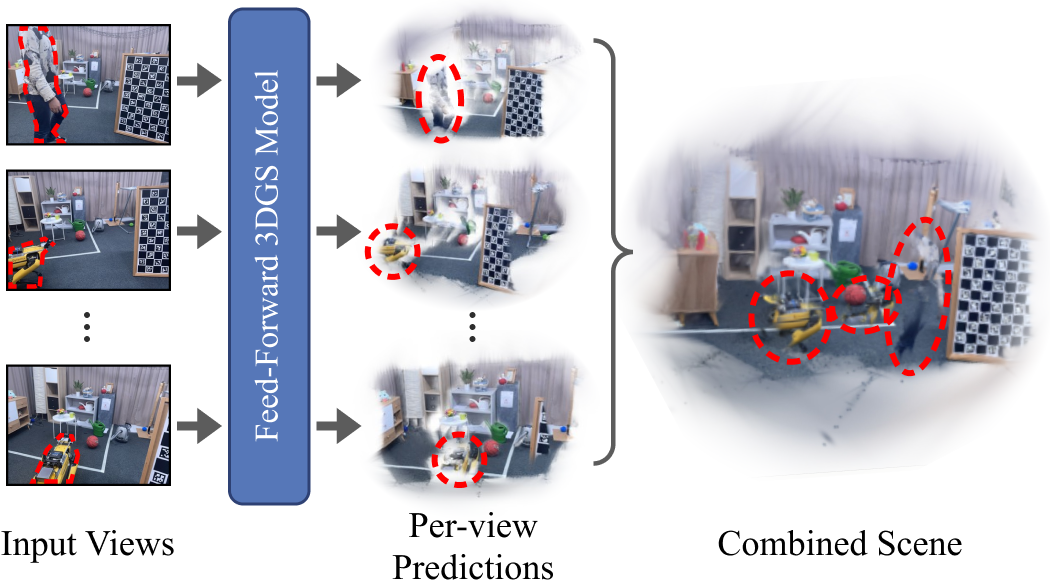}}}
    \caption{View-associated prediction in feed-forward 3DGS. Content from
individual inputs, including inconsistent distractors, can be encoded
in per-view Gaussians and remain in the combined reconstruction.}
    \label{fig:intro_analysis}
\end{figure}

Our key insight is to repurpose this per-view prediction structure as a detection cue. By excluding the Gaussians associated with one input, we expose content that is inconsistent with the remaining observations. The explicit Gaussian
representation then allows us to measure whether removing each proposed
subset improves agreement with the other inputs. Thus, the structure
that can preserve inconsistent distractors also provides the means to
identify and filter them.

Based on this insight, we propose a training-free procedure for
filtering Gaussians associated with distractors from frozen
feed-forward reconstruction models. For each input image, we compare it
with a rendering obtained after excluding its associated Gaussians.
Regions that the remaining inputs do not explain form separate removal candidates. We temporarily remove the Gaussians associated with each
candidate and evaluate their effect from selected other input views.
Only verified candidates are used to determine the final Gaussian
subsets for removal.

All operations are performed after one execution of the frozen reconstruction model; the model is neither modified nor executed again. Our method requires no additional training, learned distractor masks, or predefined distractor categories, and operates on the given inputs alone. Distractor robustness is thus added to already trained feed-forward 3DGS models as a drop-in post-processing step.

We evaluate our method on DepthSplat~\cite{depthsplat},
ReSplat~\cite{resplat}, and YoNoSplat~\cite{yonosplat} using
RobustNeRF~\cite{robustnerf} and
NeRF On-the-go~\cite{nerf-on-the-go} with varying numbers of input
views. Each + Ours result starts from the same frozen Gaussian
prediction as its corresponding baseline, isolating the effect of
filtering. We further evaluate preservation on clean scenes across four
models, including AnySplat~\cite{anysplat}, report
GenWildSplat~\cite{genwildsplat} as a trained reference, and conduct
ablation and runtime analyses.

Our contributions are as follows:
\begin{itemize}
\item We show that the native association between input views and predicted Gaussians, which allows distractors to persist in the reconstruction, can be repurposed as a cue for training-free distractor filtering.
    \item We propose a filtering procedure that forms removal candidates by excluding view-associated Gaussians and accepts only candidates verified against the other input observations, operating on a single frozen prediction.

    \item We demonstrate consistent improvements across three reconstruction models and two distractor benchmarks with varying numbers of input views, while largely preserving reconstruction quality on clean scenes across four models.
\end{itemize}
\section{Related Work}

\begin{figure*}[t]
    \centering
    {\fboxsep=0pt\fboxrule=0pt\fbox{\includegraphics[width=\dimexpr\textwidth-2\fboxrule\relax]{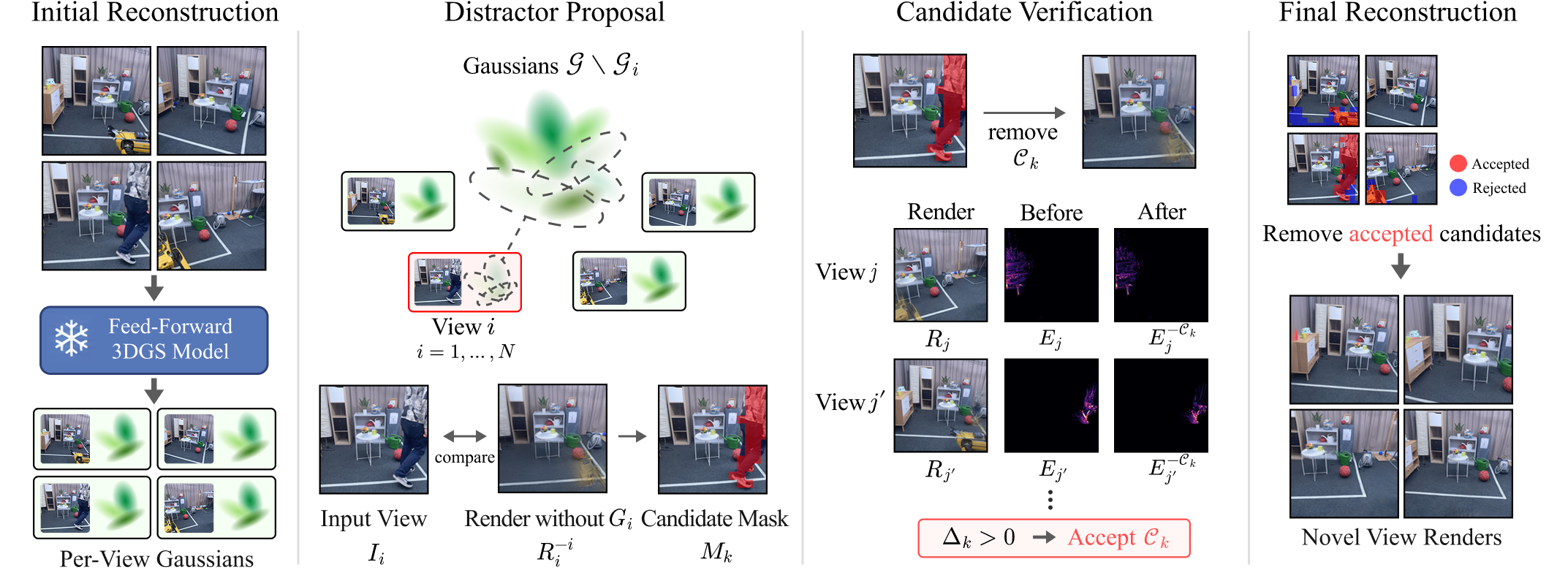}}}
    \caption{Overview of our training-free filtering procedure. For each
input, we exclude its associated Gaussian subset to form separate
candidate regions. Each candidate is first evaluated from selected
other input views, and only candidates that pass the corresponding
verification steps are removed.}
    \label{fig:method_overview}
\end{figure*}

\subsection{Feed-Forward 3D Gaussian Splatting}

Following the development of neural radiance fields
~\cite{nerf,mipnerf360}, 3D Gaussian Splatting~\cite{3dgs}
has driven rapid advances in novel view synthesis across diverse scene
settings and applications
~\cite{scaffold-gs,gaussiansplattingslam,3dgs-mcmc,4dgs,
langsplat,citygaussian,physgaussian}.
These approaches typically optimize a separate representation for
each scene. A recent direction instead develops generalizable
feed-forward models that directly predict Gaussian representations
from input images, avoiding per-scene optimization.

pixelSplat~\cite{pixelsplat} introduced feed-forward Gaussian
reconstruction from image pairs, while MVSplat~\cite{mvsplat}
extended this formulation to sparse posed views.
DepthSplat~\cite{depthsplat} further combines Gaussian reconstruction
with learned depth estimation. Subsequent methods support unposed or
otherwise less constrained inputs
~\cite{noposplat,anysplat,yonosplat}, recurrent refinement
~\cite{resplat}, token-aligned prediction~\cite{tokensplat},
voxel-aligned prediction~\cite{volsplat}, and adaptive subpixel
primitive prediction~\cite{off-the-grid}.

The reconstruction models considered in this work preserve an
association between predicted Gaussians and their input views. Their
explicit Gaussian outputs also allow selected subsets to be removed
temporarily and rendered again. We use these properties to inspect the
contribution of each input after the scene has been predicted and to
identify Gaussian subsets whose removal improves the reconstruction.

\subsection{Ignoring Distractors in 3D Reconstruction}

Distractor-free novel view synthesis has primarily been studied through
optimization-based NeRF and 3DGS pipelines. NeRF-based approaches model
transient appearance~\cite{nerf-w}, suppress inconsistent observations
with robust objectives~\cite{robustnerf}, or exploit uncertainty in
casually captured scenes~\cite{nerf-on-the-go}. Related 3D Gaussian
Splatting methods leverage pretrained features to handle inconsistent
observations~\cite{wildgaussians,spotlesssplats}, explicitly separate
static and transient representations
~\cite{desplat,hybridgs,forestsplats}, or suppress unstable artifacts
through consistency between independently optimized models
~\cite{asymgs}. Other methods derive cleaner supervision through
progressive or two-stage reconstruction~\cite{pdf-gs,dualsplat}.
These approaches identify distractors while optimizing a
scene-specific representation against its input images.

Recent work has also introduced feed-forward reconstruction models
trained specifically for distractor handling. VGTW~\cite{vgtw} builds
on VGGT and introduces distractor-aware training together with an
auxiliary mask head supervised by pixel-level annotations.
DGGS~\cite{dggs} builds on MVSplat~\cite{mvsplat} and learns mask
prediction and refinement from reference images. Its reported
inference procedure uses the predicted masks to score and reselect
references before pruning Gaussians associated with distractors.
GenWildSplat~\cite{genwildsplat}, based on
AnySplat~\cite{anysplat}, uses pretrained semantic segmentation for
transient-object masking together with curriculum learning on
synthetic and real data.

These approaches incorporate distractor handling into the reconstruction pipeline itself, relying on additional trained components or backbone-specific adaptation. Our method instead decouples distractor handling from the reconstruction model: it operates on one Gaussian prediction produced from a fixed set of inputs, with no additional training, predefined distractor categories, or learned mask prediction.
Under its reported evaluation protocol, DGGS assumes access to a scene image pool around each query view, from which reconstruction inputs are scored and reselected before reconstruction is executed again. Our method assumes only a fixed input set: the reconstruction model is executed once, and filtering operates on its resulting Gaussian representation.
\section{Method}

\begin{table*}[t]
    \centering
    {\small
    \setlength{\tabcolsep}{8pt}
    \begin{tabular}{lccccccccc}
        \toprule
        & \multicolumn{3}{c}{4 views} & \multicolumn{3}{c}{8 views} & \multicolumn{3}{c}{16 views} \\
        \cmidrule(lr){2-4} \cmidrule(lr){5-7} \cmidrule(lr){8-10}
        Method & PSNR$\uparrow$ & SSIM$\uparrow$ & LPIPS$\downarrow$ & PSNR$\uparrow$ & SSIM$\uparrow$ & LPIPS$\downarrow$ & PSNR$\uparrow$ & SSIM$\uparrow$ & LPIPS$\downarrow$ \\
        \midrule
        \multicolumn{10}{l}{\textbf{\textit{RobustNeRF dataset}}} \\
        DepthSplat & 20.10 & 0.738 & 0.267 & 19.91 & 0.706 & 0.288 & 18.44 & 0.644 & 0.369 \\
        + Ours & \textbf{21.66} & \textbf{0.762} & \textbf{0.229} & \textbf{21.37} & \textbf{0.735} & \textbf{0.242} & \textbf{19.98} & \textbf{0.690} & \textbf{0.296} \\
        \cmidrule(lr){1-10}
        YoNoSplat & 21.70 & 0.722 & 0.260 & 23.12 & 0.776 & 0.230 & 22.70 & 0.732 & 0.249 \\
        + Ours & \textbf{22.47} & \textbf{0.729} & \textbf{0.246} & \textbf{24.04} & \textbf{0.791} & \textbf{0.199} & \textbf{23.26} & \textbf{0.743} & \textbf{0.226} \\
        \cmidrule(lr){1-10}
        ReSplat & 22.24 & 0.778 & 0.270 & 22.73 & 0.763 & 0.280 & 24.71 & 0.797 & 0.238 \\
        + Ours & \textbf{22.87} & \textbf{0.782} & \textbf{0.264} & \textbf{23.12} & \textbf{0.768} & \textbf{0.273} & \textbf{24.83} & \textbf{0.799} & \textbf{0.235} \\
        \midrule
        \multicolumn{10}{l}{\textbf{\textit{NeRF On-the-go dataset}}} \\
        DepthSplat & 17.86 & 0.549 & 0.306 & 17.76 & 0.524 & 0.336 & 16.32 & 0.447 & 0.416 \\
        + Ours & \textbf{18.44} & \textbf{0.563} & \textbf{0.285} & \textbf{18.67} & \textbf{0.547} & \textbf{0.298} & \textbf{17.52} & \textbf{0.489} & \textbf{0.355} \\
        \cmidrule(lr){1-10}
        YoNoSplat & 18.35 & 0.547 & 0.329 & 18.66 & 0.561 & 0.326 & 17.95 & 0.532 & 0.350 \\
        + Ours & \textbf{19.61} & \textbf{0.577} & \textbf{0.289} & \textbf{19.64} & \textbf{0.583} & \textbf{0.295} & \textbf{19.13} & \textbf{0.564} & \textbf{0.302} \\
        \cmidrule(lr){1-10}
        ReSplat & 18.88 & 0.618 & 0.341 & 19.12 & 0.610 & 0.341 & 19.47 & 0.625 & 0.319 \\
        + Ours & \textbf{19.36} & \textbf{0.629} & \textbf{0.330} & \textbf{19.86} & \textbf{0.625} & \textbf{0.325} & \textbf{19.88} & \textbf{0.635} & \textbf{0.309} \\
        \bottomrule
    \end{tabular}
    }
    \caption{Quantitative results on the RobustNeRF and NeRF On-the-go datasets.}
    \label{tab:quantitative_main}
\end{table*}

\begin{figure*}[t]
    \centering
    {\fboxsep=0pt\fboxrule=0pt\fbox{\includegraphics[width=\dimexpr\textwidth-2\fboxrule\relax]{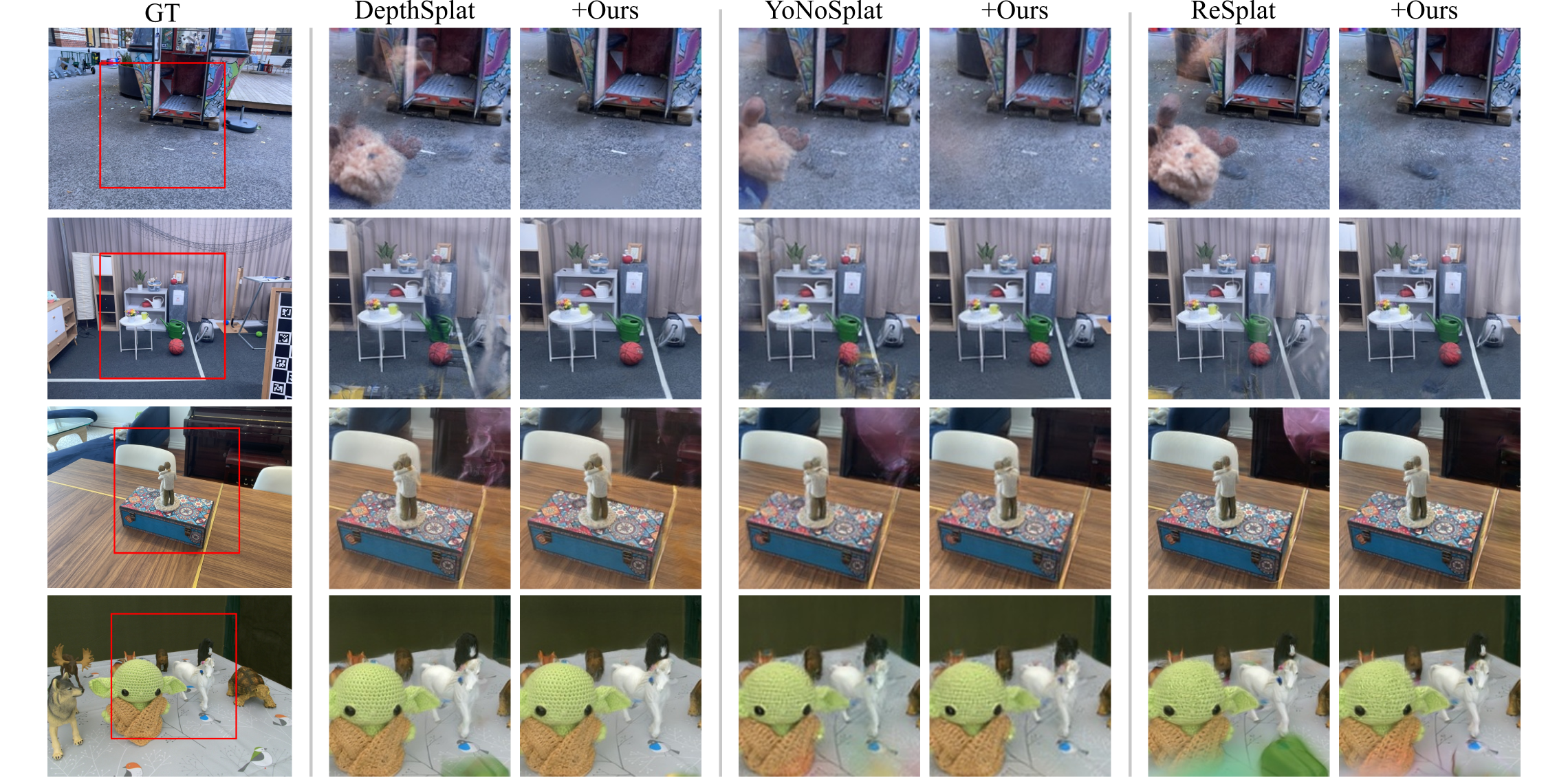}}}
    \caption{Qualitative comparisons across frozen feed-forward
reconstruction backbones. Each $+$ Ours result filters the corresponding
baseline prediction, suppressing distractor artifacts while largely
preserving surrounding scene content.}
    \label{fig:qualitative}
\end{figure*}

\subsection{Preliminaries}

\paragraph{3D Gaussian Splatting.}
3D Gaussian Splatting~\cite{3dgs} represents a scene using $N_g$
Gaussian primitives $\mathcal{G}=\{g_n\}_{n=1}^{N_g}$, each carrying
a position, covariance, opacity, and view-dependent color. We denote
by $\mathcal{R}(\mathcal{G},P)$ the image rendered from $\mathcal{G}$
under camera parameters $P$ via depth-ordered alpha compositing. Its
explicit representation allows any Gaussian subset to be removed and
the result rendered immediately.

\paragraph{Feed-Forward 3DGS.}
Given $N$ posed image-camera pairs
\begin{equation}
\mathcal{X}
=
\{(I_i,P_i)\}_{i=1}^{N},
\end{equation}
a feed-forward model $F_{\theta}$ predicts a Gaussian representation in
one network execution~\cite{pixelsplat,mvsplat,depthsplat}. We express
its output as Gaussian subsets associated with the input views:
\begin{equation}
\{\mathcal{G}_i\}_{i=1}^{N}
=
F_{\theta}(\mathcal{X}),
\qquad
\mathcal{G}
=
\bigcup_{i=1}^{N}\mathcal{G}_i.
\end{equation}
Here, $\mathcal{G}_i$ contains the Gaussian primitives associated with
input view $i$. For pixel-aligned models,
\begin{equation}
\mathcal{G}_i
=
\{g_{i,u}\mid u\in\Omega_i\},
\end{equation}
where $\Omega_i$ is the prediction domain and $u$ indexes a prediction
location. The center $\mu_{i,u}$ of $g_{i,u}$ is typically obtained by
unprojecting a predicted depth $d_{i,u}$:
\begin{equation}
\mu_{i,u}
=
\Pi^{-1}(u,d_{i,u};P_i),
\end{equation}
where $\Pi^{-1}$ denotes unprojection under camera $P_i$. More generally,
$\mathcal{G}_i$ follows the association provided by the native prediction
structure of the reconstruction model.

\subsection{Problem Setup}

We consider reconstruction from a fixed set of $N$ posed input images
that may contain transient objects visible in only a subset of the
views. At inference, no clean images, distractor masks, or additional
images for input selection are available. A frozen feed-forward model
produces $\mathcal{G}$ once, and neither the network nor the predicted
representation is optimized.

Our goal is to identify Gaussian subsets associated with distractors and to remove a subset only when the input observations themselves justify its removal.
The partition
$\{\mathcal{G}_i\}_{i=1}^{N}$ allows us to exclude temporarily the
contribution associated with each input. The explicit Gaussian
representation then allows us to render the scene before and after
removing a selected subset and measure the resulting change directly.

\subsection{Distractor Proposal}

For a quantity associated with input view $i$, the superscript $-i$
indicates that it is computed after excluding $\mathcal{G}_i$.
We render camera $P_i$ before and after excluding this subset:
\begin{equation}
R_i
=
\mathcal{R}(\mathcal{G},P_i),
\qquad
R_i^{-i}
=
\mathcal{R}(\mathcal{G}\setminus\mathcal{G}_i,P_i).
\end{equation}
Content in $I_i$ represented mainly by $\mathcal{G}_i$, including distractors observed only in view $i$, is unlikely to be reproduced in $R_i^{-i}$.

We let $\Phi$ denote the DINOv3 feature map~\cite{dinov3}, computed from each image. The feature
similarity at patch $u$ is

\begin{equation}
s_i(u)
=
\mathrm{sim}
\left(
\Phi(I_i)(u),
\Phi(R_i^{-i})(u)
\right),
\end{equation}
where $\mathrm{sim}$ denotes cosine similarity. Feature similarity is
less sensitive than direct RGB comparison to small appearance
differences between the input and rendering.

Low feature similarity may also arise from poorly reconstructed static
content or from regions that no remaining view observes.
We therefore use accumulated opacity and depth to restrict proposal
generation. Let $a_i^{-i}(u)$ denote the patch-average accumulated
opacity in $R_i^{-i}$, and $z_i(u)$ and $z_i^{-i}(u)$ the
patch-average depths before and after excluding $\mathcal{G}_i$,
computed over pixels with valid depth in both renderings; patches
without such pixels do not satisfy the depth condition.

Let $g_i(u)$ denote the support condition
$a_i^{-i}(u)>\eta$ and $z_i(u)<z_i^{-i}(u)$, with opacity threshold
$\eta$. Using similarity boundaries $\tau_1<\tau_2$, we define two
proposal ranges:
\begin{equation}
B_i^1
=
\{u\mid s_i(u)<\tau_1,\; g_i(u)\},
\end{equation}
\begin{equation}
B_i^2
=
\{u\mid \tau_1\leq s_i(u)<\tau_2,\; g_i(u)\}.
\end{equation}

The opacity condition removes regions that are left largely empty by
the remaining views. The depth condition requires exclusion of
$\mathcal{G}_i$ to reveal a farther surface, as expected when the
removed Gaussians occlude other scene content.

We extract 8-connected components independently from the two ranges
on the feature grid and upsample each component to the input image
resolution using nearest-neighbor interpolation. Components from
different ranges are not merged, even when they touch after
upsampling, and no minimum component size is imposed.

Across all input views, the resulting component masks are denoted by
$\{M_k\}_{k=1}^{L}$, and $i_k$ denotes the input view from which
$M_k$ was obtained. We use $r_k\in\{1,2\}$ to indicate whether the
component was extracted from $B_{i_k}^1$ or $B_{i_k}^2$.

Each component mask is mapped to a candidate Gaussian subset
$\mathcal{C}_k\subset\mathcal{G}_{i_k}$. For pixel-aligned models, we
use the native correspondence between prediction locations and
Gaussians. For ReSplat, we use its native rasterizer association to
select front-surface Gaussians from input $i_k$ whose projected
covariances contribute to pixels inside $M_k$.

Let $M_i^1$ and $M_i^2$ denote the unions of the component masks
obtained from the first and second similarity ranges, respectively,
for input view $i$.

\subsection{Candidate Verification}

The proposal masks identify possible distractors rather than final
removals. We first evaluate each component independently.

For input view $i$, we project the centers of $\mathcal{G}_i$ into
every other input camera. For each camera, we count the projected
centers that have positive depth and fall inside the image boundary.
The $n_v$ cameras with the largest counts form the verification set
$\mathcal{V}_i$; camera $P_i$ itself is not included.

For candidate $\mathcal{C}_k$ and each
$j\in\mathcal{V}_{i_k}$, we render the scene before and after
temporarily removing the candidate:
\begin{equation}
R_j
=
\mathcal{R}(\mathcal{G},P_j),
\qquad
R_j^{-\mathcal{C}_k}
=
\mathcal{R}(\mathcal{G}\setminus\mathcal{C}_k,P_j).
\end{equation}

The corresponding reconstruction errors are
\begin{equation}
E_j(p) = \left\|R_j(p)-I_j(p)\right\|_1,
\end{equation}
\begin{equation}
E_j^{-\mathcal{C}_k}(p) = \left\|R_j^{-\mathcal{C}_k}(p)-I_j(p)\right\|_1.
\end{equation}
The error reduction caused by removing candidate $k$ is
\begin{equation}
\delta_{j,k}(p)
=
E_j(p)-E_j^{-\mathcal{C}_k}(p).
\end{equation}
A positive value indicates that removing $\mathcal{C}_k$ reduces the
reconstruction error at pixel $p$. We weight each pixel by the
rendering change caused by the candidate:
\begin{equation}
w_{j,k}(p)
=
\left\|
R_j^{-\mathcal{C}_k}(p)-R_j(p)
\right\|_1.
\end{equation}
Proposal regions are excluded because they may contain inconsistent
content: we exclude $M_j^1$ for a candidate from the first similarity
range, and $M_j^1\cup M_j^2$ for one from the second. Let $Q_{j,k}$
denote the corresponding exclusion mask. We also ignore pixels whose
rendering change is below $\epsilon$:
\begin{equation}
\Omega_{j,k}
=
\{p\mid
p\notin Q_{j,k},\;
w_{j,k}(p)>\epsilon\}.
\end{equation}

We pool all valid pixels from the selected views into one verification score:
\begin{equation}
\Delta_k
=
\frac{
\sum_{j\in\mathcal{V}_{i_k}}
\sum_{p\in\Omega_{j,k}}
w_{j,k}(p)\,\delta_{j,k}(p)
}{
\sum_{j\in\mathcal{V}_{i_k}}
\sum_{p\in\Omega_{j,k}}
w_{j,k}(p)
}.
\end{equation}
A positive score indicates that removing $\mathcal{C}_k$ improves
agreement with the selected input observations outside the proposal
regions. If the denominator is zero, we set $\Delta_k=0$.

Candidates with positive individual scores proceed to final Gaussian
selection. We dilate their component masks at the input image
resolution and intersect the expanded masks again with the support
condition $g_{i_k}$ used during proposal generation. Each resulting
mask is mapped to a subset
$\widehat{\mathcal{C}}_k\subset\mathcal{G}_{i_k}$ using the same
model-specific association as above.

Candidates from the first similarity range, whose lower similarity
already indicates a clear mismatch, require no further test.
Candidates from the second range carry weaker evidence, and we verify
them further. For each such candidate, we additionally render every
input camera after excluding the Gaussian subset associated with that
camera:
\begin{equation}
R_j^{-j}
=
\mathcal{R}(\mathcal{G}\setminus\mathcal{G}_j,P_j),
\end{equation}
and after additionally removing the expanded candidate:
\begin{equation}
R_j^{-j,-\widehat{\mathcal{C}}_k}
=
\mathcal{R}
\left(
\mathcal{G}\setminus
\left(
\mathcal{G}_j\cup\widehat{\mathcal{C}}_k
\right),
P_j
\right).
\end{equation}
Using the same rendering-change-weighted error reduction as above, we
aggregate all valid pixels over all input views while excluding
$M_j^1\cup M_j^2$. The candidate proceeds only when this additional
score is positive.

Finally, the second-range candidates that pass both individual tests
are evaluated together: starting from the representation with the
accepted first-range candidates removed, we additionally remove all
remaining second-range candidates and aggregate the same
rendering-change-weighted error reduction over all input views outside
$M_j^1\cup M_j^2$. They are retained when this score is positive and
restored together otherwise; first-range candidates are unaffected.

\subsection{Final Reconstruction}

Let $\mathcal{A}_1$ denote candidates from the first similarity range
with positive individual scores, and $\mathcal{A}_2$ those from the
second range that pass the individual and additional steps. If their
combined verification fails, we set $\mathcal{A}_2=\emptyset$.

The filtered Gaussian representation is
\begin{equation}
\mathcal{G}_{\mathrm{out}}
=
\mathcal{G}
\setminus
\bigcup_{k\in\mathcal{A}_1\cup\mathcal{A}_2}
\widehat{\mathcal{C}}_k.
\end{equation}
In implementation, selected Gaussians are removed by setting their
opacity to zero; all others remain unchanged.

\section{Experiments}

\paragraph{Experimental Setup.}
We evaluate on the RobustNeRF~\cite{robustnerf} and NeRF On-the-go~\cite{nerf-on-the-go} benchmarks using 4, 8, and 16 input views containing distractors, covering sparse to moderately dense capture settings. For each scene and input count, we construct four view configurations and average the results over them. Each configuration starts from an input view and a test view sharing
high COLMAP~\cite{colmap} sparse-point visibility, and views are added to increase
the scene coverage jointly observed by both sets.

We report PSNR, SSIM~\cite{ssim}, and LPIPS~\cite{lpips} on the test
images. Each reconstruction model is evaluated at the resolution used
by its released inference configuration.
We use $\tau_1=0.50$, $\tau_2=0.70$, $n_v=3$, and $\eta=0.50$ for all
datasets. Components are extracted with 8-connectivity without a
minimum-size threshold. Candidates that pass individual verification
are dilated by 9 pixels before their final Gaussian subsets are
determined. The additional verification for the second similarity range aggregates all input views. We use $\epsilon=0.002$ and accept a verification
result only when its score is positive. These settings are fixed
across all datasets, input counts, and reconstruction models. No
training, fine-tuning, or scene-specific optimization is performed.
Experiments were run on NVIDIA RTX 3090 and
RTX 4090 GPUs. All runtime measurements are obtained
on a single RTX 4090.

\begin{table}[!t]
    \centering
        \footnotesize
        \begin{tabular}{lccc}
            \toprule
            Variant & PSNR$\uparrow$ & SSIM$\uparrow$ & LPIPS$\downarrow$ \\
            \midrule
            Full Method & 20.42 & 0.657 & 0.279 \\
            RGB Difference & 20.31 & 0.654 & 0.284 \\
            Single Threshold ($<\tau_2$) & 20.10 & 0.649 & 0.291 \\
            Single Threshold ($<\tau_1$) & 20.32 & 0.655 & 0.282 \\
            Without Verification & 20.02 & 0.645 & 0.296 \\
            Vanilla & 19.56 & 0.641 & 0.301 \\
            \bottomrule
        \end{tabular}
        \caption{Ablation study.}
        \label{tab:ablation}
\end{table}

\begin{table}[!t]
    \centering
    \small
    \setlength{\tabcolsep}{1.8pt}
    \begin{tabular}{lcccccc}
        \toprule
        & \multicolumn{3}{c}{4 views} & \multicolumn{3}{c}{6 views} \\
        \cmidrule(lr){2-4} \cmidrule(lr){5-7}
        Method & PSNR$\uparrow$ & SSIM$\uparrow$ & LPIPS$\downarrow$ & PSNR$\uparrow$ & SSIM$\uparrow$ & LPIPS$\downarrow$ \\
        \midrule
        \multicolumn{7}{l}{\textbf{\textit{RobustNeRF dataset}}} \\
        AnySplat & 13.97 & 0.457 & 0.492 & 15.78 & 0.506 & 0.438 \\
        + Ours & 14.48 & 0.469 & 0.468 & 16.63 & 0.526 & 0.400 \\
        GenWildSplat & 13.99 & 0.484 & 0.516 & 14.39 & 0.482 & 0.532 \\
        \midrule
        \multicolumn{7}{l}{\textbf{\textit{NeRF On-the-go dataset}}} \\
        AnySplat & 13.66 & 0.287 & 0.469 & 14.96 & 0.328 & 0.437 \\
        + Ours & 13.96 & 0.293 & 0.447 & 15.38 & 0.336 & 0.411 \\
        GenWildSplat & 13.50 & 0.315 & 0.544 & 13.73 & 0.315 & 0.546 \\
        \bottomrule
    \end{tabular}
\caption{Comparison with feed-forward models in the four- and
six-view settings. Our method filters the frozen AnySplat prediction;
GenWildSplat uses its released model.}
    \label{tab:comparison_existing_methods}
\end{table}

\paragraph{Reconstruction Models and Comparisons.}
We apply our method to the released DepthSplat~\cite{depthsplat},
ReSplat~\cite{resplat}, and YoNoSplat~\cite{yonosplat} models. All
three are evaluated with input camera poses, which removes pose
estimation as a confounding factor. ReSplat uses its 8-view checkpoint
for up to 8 inputs and its 16-view checkpoint for 16.
Each baseline result and its corresponding + Ours result use
identical images, cameras, model weights, and initial Gaussian
prediction; only the selected Gaussian subsets differ, isolating the
effect of filtering. We additionally evaluate
AnySplat~\cite{anysplat} with and without our method, using its
estimated cameras for proposal generation and verification, and
include the released GenWildSplat~\cite{genwildsplat} model as an
AnySplat-based reference. GenWildSplat targets unconstrained image
collections and uses semantic masks for predefined transient object
categories. Since it reports reconstruction with two to six input
views, we use four- and six-view settings. DGGS~\cite{dggs} is
excluded, as no implementation or weights were publicly available at
submission and its inference requires additional scene images beyond
the given inputs.

\paragraph{Quantitative Results.}
Table~\ref{tab:quantitative_main} compares each reconstruction before and after filtering. Our method improves PSNR, SSIM, and LPIPS across all evaluated reconstruction models, datasets, and input counts. Since each pair shares the same images, model weights, and Gaussian prediction, these gains are attributable to filtering alone.

\begin{table}[!t]
    \centering
    \small
    \setlength{\tabcolsep}{1.8pt}
    \begin{tabular}{lcccccc}
        \toprule
        & \multicolumn{3}{c}{4 views} & \multicolumn{3}{c}{6 views} \\
        \cmidrule(lr){2-4} \cmidrule(lr){5-7}
        Method & PSNR$\uparrow$ & SSIM$\uparrow$ & LPIPS$\downarrow$ & PSNR$\uparrow$ & SSIM$\uparrow$ & LPIPS$\downarrow$ \\
        \midrule
        \multicolumn{7}{l}{\textbf{\textit{With GT Pose}}} \\
        DepthSplat & 23.14 & 0.804 & 0.169 & 22.04 & 0.769 & 0.197 \\
        + Ours & 23.14 & 0.804 & 0.170 & 22.04 & 0.769 & 0.197 \\
        \cmidrule(lr){1-7}
        YoNoSplat & 24.98 & 0.802 & 0.167 & 25.24 & 0.813 & 0.163 \\
        + Ours & 24.98 & 0.802 & 0.169 & 25.17 & 0.811 & 0.165 \\
        \cmidrule(lr){1-7}
        ReSplat & 24.88 & 0.809 & 0.224 & 24.49 & 0.799 & 0.232 \\
        + Ours & 24.83 & 0.809 & 0.224 & 24.49 & 0.799 & 0.232 \\
        \midrule
        \multicolumn{7}{l}{\textbf{\textit{With Estimated Pose}}} \\
        AnySplat & 16.40 & 0.506 & 0.412 & 17.69 & 0.558 & 0.348 \\
        + Ours & 16.39 & 0.506 & 0.413 & 17.69 & 0.558 & 0.349 \\
        GenWildSplat & 14.58 & 0.507 & 0.483 & 14.81 & 0.506 & 0.487 \\
        \bottomrule
    \end{tabular}
    \caption{Quantitative results on clean scenes.}
    \label{tab:clean_scene}
\end{table}

\begin{figure}[!t]
    \centering
    \includegraphics[width=\linewidth]{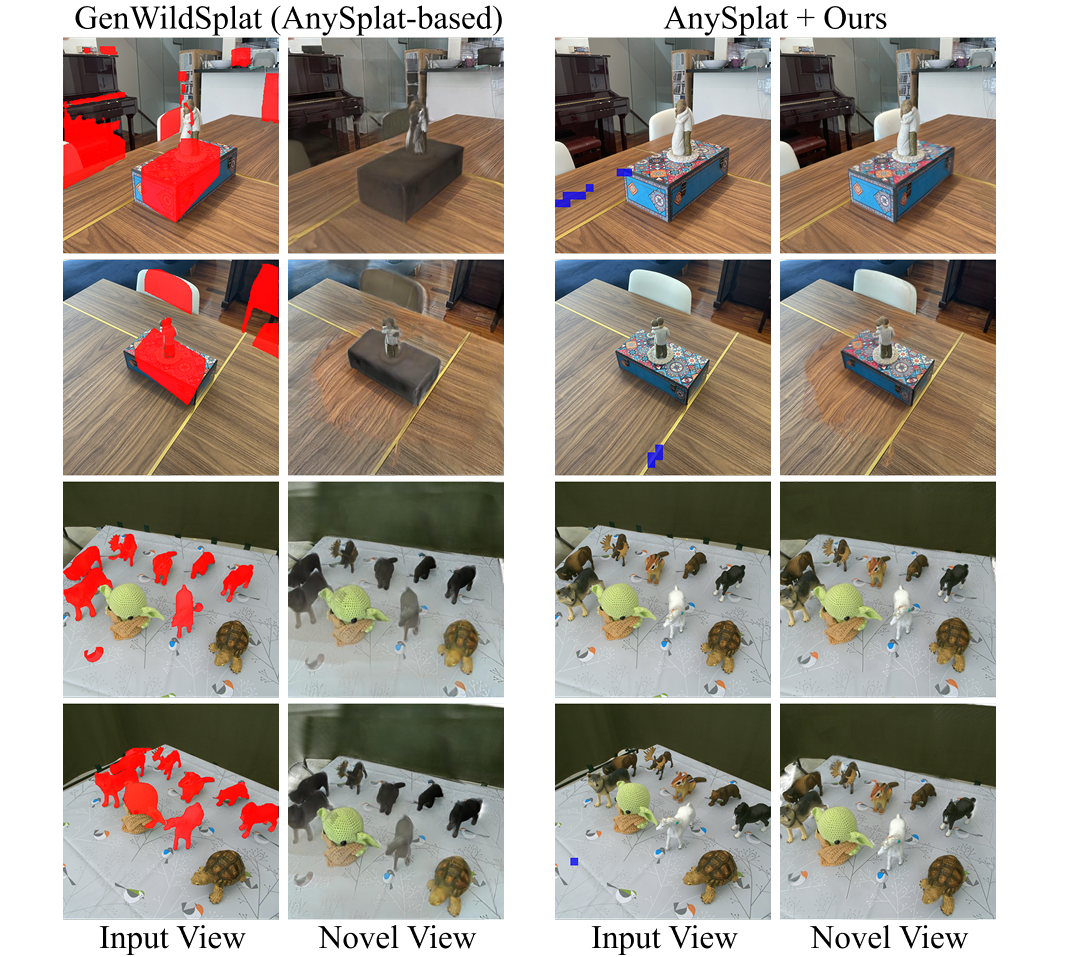}
    \caption{Qualitative results on clean scenes. For our method,
    proposal regions (blue) are rejected during verification, leaving the
    reconstruction unchanged, whereas GenWildSplat's semantic masking (red)
    can remove static objects.}
    \label{fig:clean_scene}
\end{figure}

\paragraph{Qualitative Results.}
Figure~\ref{fig:qualitative} shows that transient people and objects
can produce blurred, duplicated, or floating artifacts in the original
reconstructions. Our method suppresses these artifacts across multiple
backbones while largely preserving nearby static scene content. All
methods are compared using the same enlarged image regions.

\paragraph{Ablation Studies.}
Table~\ref{tab:ablation} reports averages over the ten scenes from both datasets and three backbones in 4-view setting. Removing verification causes the largest degradation, confirming proposal regions should not be removed directly. Replacing DINOv3 similarity with a simple RGB difference retains most of the improvement over the vanilla reconstruction; DINOv3 and dual proposal ranges add further gains.

\paragraph{Comparison with In-the-Wild Feed-Forward Model.}Table~\ref{tab:comparison_existing_methods} compares AnySplat,
AnySplat with our method, and GenWildSplat on the two distractor
benchmarks using four and six input views. Here all methods operate
from the cameras estimated by AnySplat, which takes unposed images as
input; our proposal generation and verification use the same cameras.
Our method improves the frozen AnySplat prediction across both
datasets and input settings, and achieves higher average performance
than the released GenWildSplat model under this protocol. GenWildSplat
builds on AnySplat and uses semantic masks for predefined transient
categories, making it the closest trained counterpart to ours.

\begin{table}[!t]
    \centering
    \small
    \setlength{\tabcolsep}{2pt}
    \begin{tabular*}{\columnwidth}{@{\extracolsep{\fill}}lcccc}
        \toprule
        Method & Views & Inference & Proposal & Verification \\
        \midrule
        ReSplat & 4 & 0.22 & 0.23 & 0.44 \\
        & 8 & 0.38 & 0.35 & 1.10 \\
        & 16 & 0.78 & 0.64 & 3.22 \\
        \cmidrule(lr){1-5}
        DepthSplat & 4 & 0.09 & 0.19 & 0.58 \\
        & 8 & 0.14 & 0.42 & 2.82 \\
        & 16 & 0.28 & 1.02 & 16.95 \\
        \cmidrule(lr){1-5}
        YoNoSplat & 4 & 0.22 & 0.12 & 0.52 \\
        & 8 & 0.25 & 0.23 & 2.83 \\
        & 16 & 0.39 & 0.59 & 14.77 \\
        \bottomrule
    \end{tabular*}
    \caption{Mean processing time in seconds, averaged over the
RobustNeRF and NeRF On-the-go scenes.}
    \label{tab:overhead_analysis}
\end{table}

\paragraph{Preservation on Clean Scenes.}
Table~\ref{tab:clean_scene} and Figure~\ref{fig:clean_scene} evaluate the four clean RobustNeRF scenes under posed and estimated-pose settings. Across four
reconstruction models, our method leaves the original reconstructions
nearly unchanged: proposal regions on clean inputs are often rejected
during verification, as shown in Figure~\ref{fig:clean_scene}.
GenWildSplat instead removes objects by semantic category, so static
objects in its predefined transient categories can also be
masked.

\paragraph{Overhead Analysis.}
Table~\ref{tab:overhead_analysis} reports the reconstruction model
execution time and the additional time of our filtering procedure for
each input count. Filtering runs once per scene, after the Gaussian
representation has been predicted, and produces a single filtered
representation $\mathcal{G}_{\mathrm{out}}$; rendering a novel view
requires no input reselection, re-reconstruction, or other per-view
processing. Verification dominates the filtering time, as it renders
each candidate from multiple views, and grows with the number of views
and candidates.

\section{Conclusion}

We showed that the per-view prediction structure of feed-forward 3DGS
models enables training-free distractor filtering: excluding the
Gaussian subset associated with each input exposes content that is
inconsistent with the remaining observations, and verification by
rendering retains only candidates whose removal improves
reconstruction. Across multiple reconstruction models and distractor
benchmarks, the resulting procedure consistently reduces distractor
artifacts while preserving quality on clean scenes, without retraining
or scene-specific optimization. Distractor robustness can thus be
obtained from the reconstruction itself, with the model left
untouched.
Verification cost grows with the number of input views and candidates,
since each candidate is checked by rendering. The individual checks
are independent, leaving room for batched or parallel evaluation in
larger collections. Our method also assumes an association between
predicted Gaussians and input views, which the reconstruction models
considered here natively provide; extending the procedure to
architectures without this structure is an open direction. Future work may further extend the framework with object-level
reasoning and geometry-aware generative completion, enabling more
structured filtering and reconstruction of newly exposed regions.

\bibliography{aaai2027}


\end{document}